\documentclass[10pt]{article}
\date{}
\usepackage[left=2.2cm,right=2.2cm,top=2.5cm,bottom=2.5cm]{geometry}

\usepackage{newfloat}
\usepackage{graphicx}
\usepackage{graphics}
\usepackage{xcolor}
\usepackage{multirow}
\usepackage{numprint}
\usepackage{amsmath}
\usepackage{amsthm}
\usepackage{soul}
\usepackage{threeparttable}
\usepackage{booktabs}
\usepackage{authblk}
\usepackage[font=bf,size=small]{caption}
\usepackage[font=normalfont]{subcaption}
\npthousandsep{,}\npthousandthpartsep{}

\theoremstyle{definition}

\newenvironment{packed_enum}{
	\begin{enumerate}
		\setlength{\itemsep}{0pt}
		\setlength{\parskip}{0pt}
		\setlength{\parsep}{0pt}
	}
	{\end{enumerate}}

\makeatletter
\g@addto@macro\normalsize{%
	\setlength\abovedisplayskip{1pt}
	\setlength\belowdisplayskip{1pt}
	\setlength\abovedisplayshortskip{1pt}
	\setlength\belowdisplayshortskip{1pt}
}
\makeatother

\newcommand\blfootnote[1]{%
  \begingroup
  \renewcommand\thefootnote{}\footnote{#1}%
  \addtocounter{footnote}{-1}%
  \endgroup
}

\title{Compositional Generalization in Spoken Language Understanding}

\author{Avik Ray$^1$ \thanks{avik@utexas.edu}}
\author{Yilin Shen$^2$ \thanks{yilin.shen@samsung.com}}
\author{Hongxia Jin$^2$ \thanks{hongxia.jin@samsung.com}}
\affil{{Department of Electrical and Computer Engineering} \authorcr
{$^1$ Amazon Alexa, USA} \authorcr
{$^2$} Samsung Research America, USA}

\begin{document}

\maketitle
 
\begin{abstract}
State-of-the-art spoken language understanding (SLU) models have shown tremendous success in benchmark SLU datasets, yet they still fail in many practical scenario due to the lack of model compositionality when trained on limited training data.
In this paper, we study two types of compositionality: \emph{novel slot combination}, and \emph{length generalization}.
We first conduct in-depth analysis, and find that state-of-the-art SLU models often learn spurious slot correlations during training, which leads to poor performance in both compositional cases.
To mitigate these limitations, we create \emph{the first compositional splits} of benchmark SLU datasets and we propose \emph{the first compositional SLU model}, including compositional loss and paired training that tackle each compositional case respectively.
On both benchmark and compositional splits in ATIS and SNIPS, we show that our compositional SLU model significantly outperforms (up to $5\%$ F1 score) state-of-the-art BERT SLU model. 
\end{abstract}
\noindent\textbf{Index Terms}: spoken language understanding, compositional generalization\blfootnote{This work was completed when author $1$ was at Samsung Research America.}

\section{Introduction} \label{sec:intro}

Spoken language understanding is an important component of task-oriented dialog systems powering today's voice controlled AI agents, and chat bots. Intent classification and slot tagging are two main sub-tasks in SLU \cite{HakTurCelChenGao:16,LiuLane:16,KimLeeStratos:17,GooGaoHsuHuo:18,WangShenJin:18}. Human language is inherently compositional \cite{chomsky1957mouton}, and humans possess the ability to understand infinite new utterances by focusing on relevant informative sub-parts of the utterance which were learned previously \cite{fodor1988connectionism}. In this work, we consider informative sub-parts of an utterance containing slots. For example, humans can understand {\bf slot value} ``boston'' is of the {\bf slot type/label} {\em B-to-city} from the utterance {\em ``show flights to \underline{boston}''}. Similarly, from another utterance {\em ``find flights from \underline{atlanta}''}, they can learn ``atlanta'' has the slot type {\em B-from-city}. If presented with a new utterance {\em ``show flights from \underline{atlanta} to \underline{boston}''}, humans can still infer the correct slot labels of ``atlanta'' and ``boston'', even though they have never seen these slots appear together before in the same utterance. Despite their success, current state-of-the-art SLU models \cite{BERTSLUChenZhuoWanh:19}, based on pre-trained language models \cite{BERTDevlinCLT:19,Roberta:19}, struggle to perform such simple compositional generalization for slot tagging, as we demonstrate in this work.

\begin{figure}[t]
	\centering
	\includegraphics[height=3in]{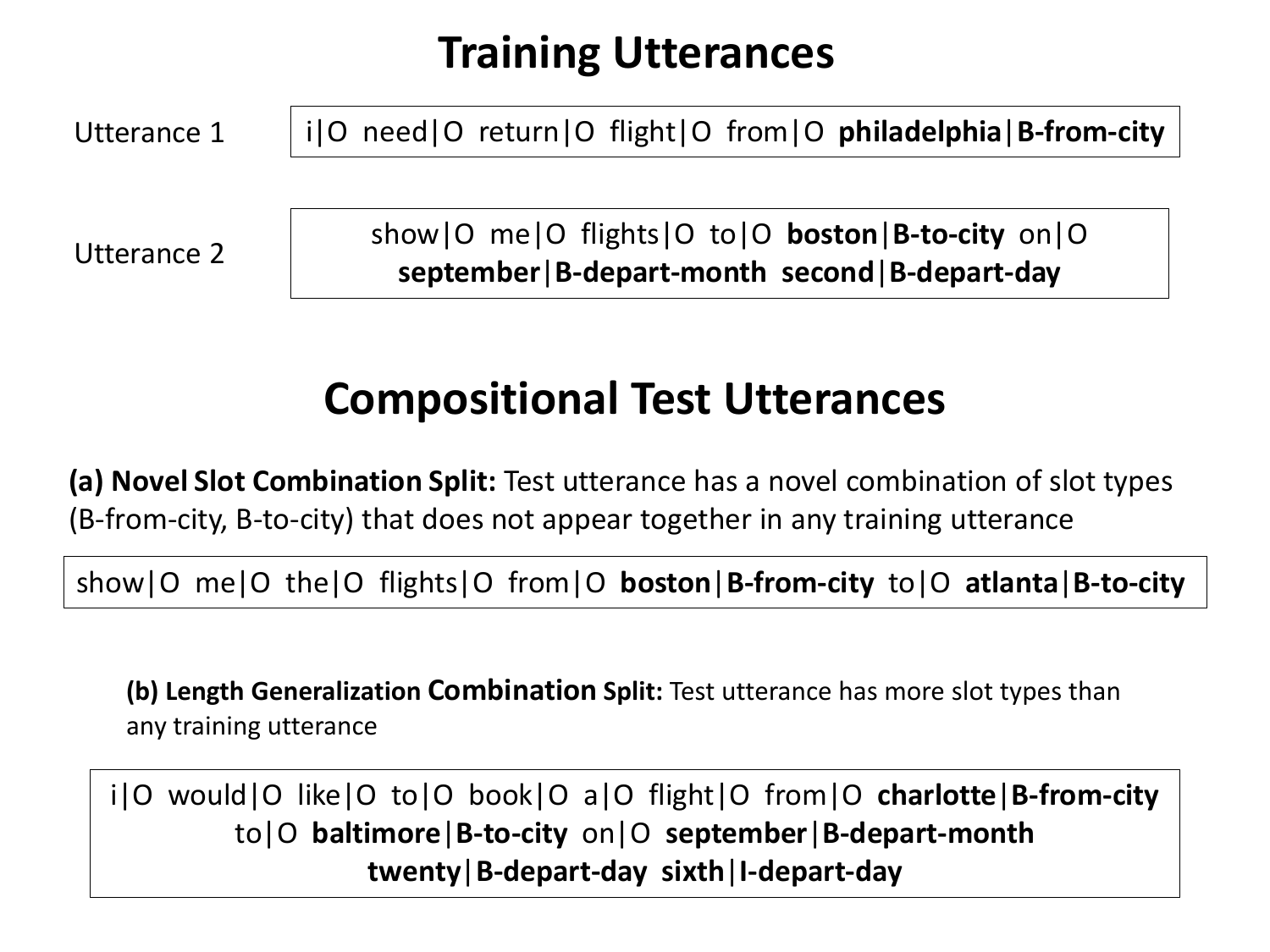}
	\caption{Example Utterances in Two Types of Slot Compositionalities}
	\label{fig:qualitative}
\end{figure}

In this paper, we investigate two main aspects of slot compositionality; (a) identifying a {\bf novel combination of slot types} in an utterance which was never seen during training, and (b) identifying more number of slot types per utterance than any training utterance, which we refer to as {\bf length generalization}. It is vitally important for SLU models to perform both these types of compositional generalization due to the following reasons. Firstly, in domains with a large number of slots (e.g. airline reservation), it can be both time-consuming and expensive to collect and annotate training utterances corresponding to each possible combination of slots. Secondly, for resource constrained cold-start skill developers \cite{ShenRPJ:18}, it is cheaper and easier to annotate a small number of short utterances (with just one or two slots) for training, than longer utterances with many slots which the SLU model may encounter after deployment. Building compositional SLU models which can generalize well under both these settings is vital for both scalable development, and reliability of future AI agents.

Due to a lack of compositional objective during training, existing SLU models fail to learn the correct dependence of slot words on the relevant informative words that convey their meaning. Instead, they often rely on spurious slot correlations to make their decision. When these models encounter an utterance with a novel combination of slots unseen during training, they fail to exploit this learned correlation, hence do not generalize. When the models encounter longer utterances with many slots per utterance, than they have seen during training, they often fail due to poor quality slot representations under a longer sentence context. While some techniques have been proposed to improve compositionality of sequence-to-sequence models in small synthetic datasets \cite{Lake:19,GordonLBB:20,NyeS0L:20,ChenLYSZ:20}, these are computationally expensive to train, and hard to scale on real-world datasets. 

\noindent{\bf Main contributions:} In this work, we improve compositional generalization of SLU models by using explicit compositional objectives during training, and develop novel data augmentation technique that helps generalization to longer utterances. Our proposed methods are practical and enable SLU models to scale to large real-world datasets. Our main contributions are:

\begin{packed_enum}
    \item We create the first compositional splits of benchmark SLU datasets (ATIS, and SNIPS). These splits can be used as a new benchmark to evaluate compositional generalization properties of SLU models.
    
    \item We explore two types of slot compositional generalization, {\em novel slot combination}, and {\em length generalization}, and conduct in-depth analysis to investigate why existing SLU models perform poorly in both cases.
    
    \item We propose a new compositional loss that improves compositionality of SLU models to utterances with unseen slot combinations, and a new paired training technique that improves length generalization of SLU models.
    
    \item We show that our new compositional SLU model can achieve significant (up to $5\%$) improvement in slot tagging F1 score on our new compositional splits.
    
\end{packed_enum}

\section{Compositional Benchmark Datasets} \label{sec:dataset}

In this section, we propose our method to create compositional splits of benchmark SLU datasets.

In order to systematically evaluate compositional generalization of SLU models, we start with two benchmark SLU datasets. The first benchmark dataset {\bf ATIS} \cite{DahlBBFHPPRS94} contains utterances related to airline reservation. We consider the data split from \cite{HeY:05,HakTurCelChenGao:16} containing \numprint{4978} training , and \numprint{893} test utterances in the standard split $(T_{\operatorname{train}},T_{\operatorname{test}})$. We also use the second benchmark dataset \textbf{SNIPS} \cite{SNIPS:18} containing various utterances in entertainment, weather, and restaurant domains. In the standard split SNIPS has \numprint{13784} training, and \numprint{700} test utterances. 
For each dataset, we create two compositional train/test splits by selecting a subset of utterances from their standard train/test splits. 

\subsection{Novel Slot Combination Split:}

Human can easily identify a slot type in isolation just by focusing on most informative words which are used to describe a slot, even when presented with an utterance having a new combination of two or more slots types that were not seen during learning (training) phase (also referred as {\em systematicity} in cognitive science \cite{fodor1988connectionism}). To test this aspect of compositional generalization, we create a train/test split where none of the combination (or set) of slot types present in a test utterance appear during training. We generate this split (referred as {\em novel slot combination}) using the following steps: (a) We remove from standard training set all utterances which have a combination of slot types that appear in the standard test set. We do not remove utterances with a single slot since these are fundamental examples from which the model learns the true semantic meaning of such slots. (b) In order to better separate compositional generalization with OOV generalization \cite{RayShenJin:19,YanHXLMHX:20}, we replace any OOV slot values with a randomly selected slot value (but of the same slot label) from the training set to generate the final test set. Figure \ref{fig:qualitative} shows example utterances from our novel slot combination split of ATIS dataset.  

\subsection{Length Generalization Split:}

Sequence based neural network models are inherently poor at generalization to longer sequences than what it observed during training \cite{SCANLakeB:18,GSCANRuisABBL:20,Hupkes:20}. Note that, the informativeness of an utterance directly depends on the number of slots present in the utterance, but it does not necessarily depend on actual length of the utterance. 
Hence, to test length generalization we consider the number of slots in the utterance to generate the split. Increasing the number of slots in an utterance also naturally increases its length. We create compositional train/test splits to test length generalization as follows: (a) From the standard training set, we only select utterances which have number of slots less than or equal to a fixed integer $k.$ (we use $k=2$ in our experiments) (b) We also remove from the test set utterances with slot combinations in the training set and substitute OOV slot values as before. We test if an SLU model has the ability to identify slots when number of slots in the utterance can be much larger than that observed during training. Figure \ref{fig:qualitative} shows an example utterance from our ATIS length generalization split. Table \ref{tab:comp_results} reports the split sizes.

\section{Our Method} \label{sec:model}

In this section, first we describe the baseline SLU model which we consider in our experiments, and explore why they have poor compositionality. Later, we propose new techniques to improve compositional generalization of SLU models. We use the following notations: Let $\mathbf{x} = (x_1, \hdots, x_n)$ denote an input utterance, where words/tokens $x_i \in \mathcal{V},$ the vocabulary. Each input token $x_i$ is annotated by a slot label $y_i \in \mathcal{Y},$ the slot vocabulary. We consider slot labels in the standard IOB format, where label `O' denotes the word/token that does not belong to any slot.

\subsection{Baseline SLU Model Analysis} \label{sec:bert_baseline}

Large pre-trained language models have been shown to be successful in most natural language understanding tasks. Our baseline SLU model is based on one such model BERT \cite{BERTDevlinCLT:19}, which also achieves state-of-the-art on benchmark SLU datasets. Our model is similar to the implementation in \cite{BERTSLUChenZhuoWanh:19}. We train the model jointly on intent classification and slot tagging tasks using the objective: $\mathcal{L} = \mathcal{L}_{\operatorname{intent}} + \lambda_1 \mathcal{L}_{\operatorname{slot}},$ where $\mathcal{L}_{\operatorname{intent}}$ is the intent classification loss, $\mathcal{L}_{\operatorname{slot}}$ is the slot tagging loss, and $\lambda_1$ is a hyper-parameter to balance the losses.

\noindent{\bf Key issue:} Although the BERT baseline model can achieve SOTA on standard splits of benchmark ATIS and SNIPS datasets, we observe that it often suffers significant drop in slot tagging performance on our compositional splits. Recall that in BERT, the self-attention layer of the transformer computes the attention distribution for each attention head $h$ as follows:
\begin{equation}
P^h = \operatorname{Softmax}\left( \frac{1}{\sqrt{d}}HW_h^Q (HW_h^K)^T \right)  \label{eq:self_attn}
\end{equation}
where $W_h^Q, W_h^K$ are the query, and key projection matrices of the $h$-th attention head, $H$ is the output hidden layer vectors of previous layer, and $d$ is the head dimension. By plotting this attention head distribution we can observe how much information each token contributes to the final slot label output logit. A human identifies a slot type by focusing on the surrounding most informative words in the utterance that help to convey the semantic meaning of the slot. One may expect that the final transformer layer in SLU model performs the same by providing higher attention weights to informative words corresponding to a slot value. However, we observe that this is not always the case. For example, as shown in Figure \ref{fig:attn_map_comparison} in utterance {\em ``play rock from the eighties''}, in order to identify the slot value {\em ``rock''}, the BERT SLU model gives more attention to a different slot {\em ``eighties''}, than informative context word ``play''. This could indicate that BERT SLU model often learns spurious correlations among context and slot words. Therefore, when such slots appear in an utterance with a new combination of slot not seen during training, the BERT model may fail to identify them. This results in poor slot tagging performance of BERT SLU model in our compositional splits (Section \ref{sec:experiment}).

\subsection{Our Compositional SLU models} \label{sec:comp_model}

We now describe two main techniques that we show can improve compositional generalization of SLU models.

\subsubsection{New Compositional Loss to Improve Slot Combination Compositionality:}

\begin{figure}[t]
	\centering
	\includegraphics[width=0.5\textwidth]{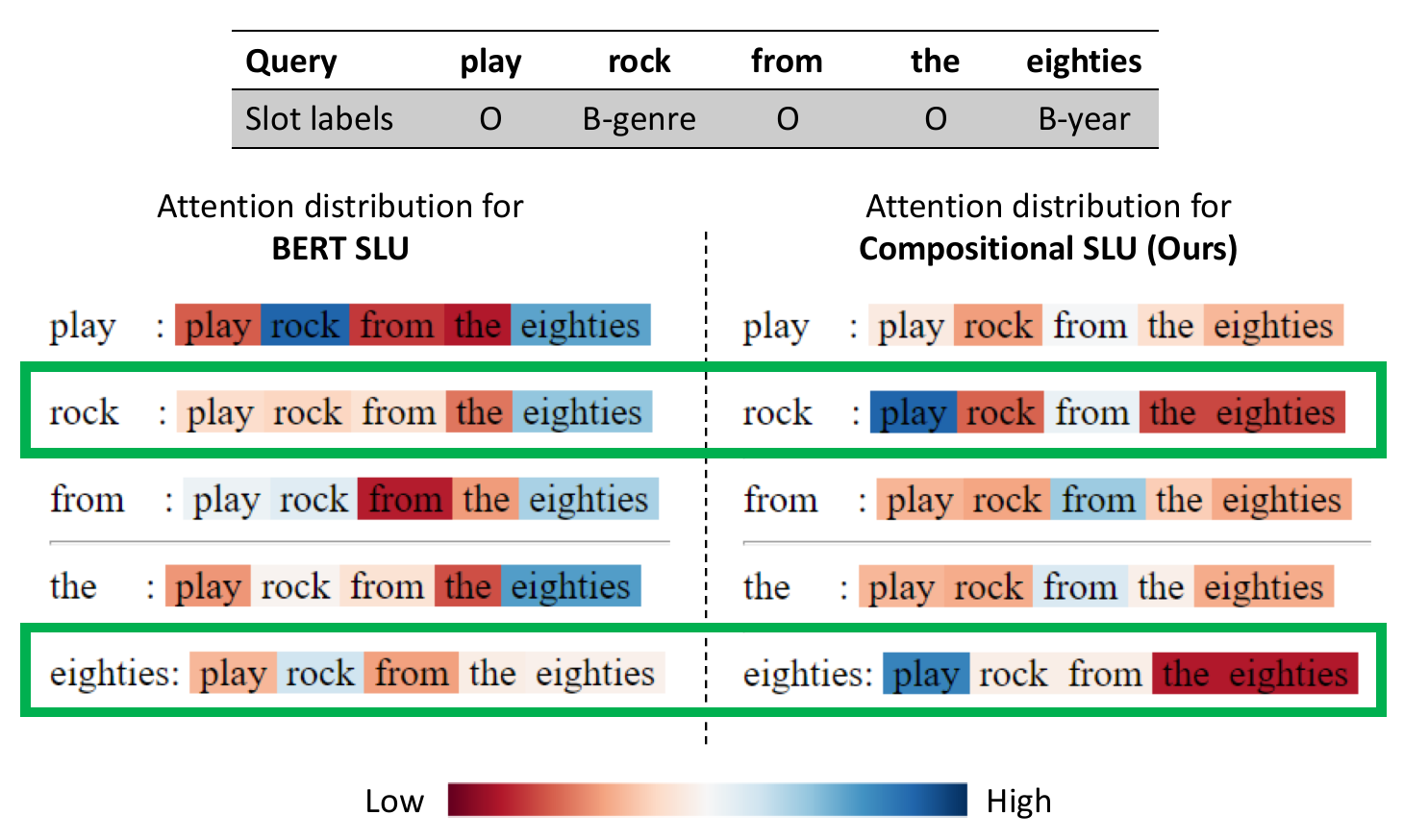}
	\caption{Visualization of the attention map (averaged over heads) for SNIPS utterance using baseline BERT SLU model (left), and our Compositional SLU model (right). {\em Blue} indicates high attention weights, and {\em Red} indicates low weights. {\em Green box} highlights the attention distribution corresponding to slot words. Compositional SLU model focus more on informative words to infer slot labels, and reduces spurious slot correlations, compared to BERT SLU model.}
	\label{fig:attn_map_comparison}
\end{figure}

We develop a new compositional loss which reduces spurious slot correlation and encourages the SLU model to focus its attention on informative context words. Intuitively, if two words have different slot labels, they should be identifiable based on a disjoint set of words. For example, for the utterance {\em ``play rock from the eighties''} (Figure \ref{fig:attn_map_comparison}), to identify slot label for the word {\em ``rock''} it is sufficient to focus on context words $ S_{\operatorname{rock}}=\{play, rock\}.$ To identify slot label for word {\em ``eighties''} it is sufficient to focus on words $ S_{\operatorname{eighties}}=\{play, rock, from, eighties\}.$ Note that these two sets of words are different. Our compositional loss is a sum of two loss functions as follows:
\begin{eqnarray}
\mathcal{L}_{\operatorname{slot-pair}} &=& \frac{1}{N_1} \sum_{h} \sum_{i,j: y_i \neq y_j \neq O} \operatorname{KL}(P^h_i,P^h_j) \label{eq:slot_pair_obj} \\
\mathcal{L}_{\operatorname{non-deg}} &=& \frac{1}{N_2} \sum_{h} \sum_{i: y_i \neq O} \operatorname{KL}(P^h_i, \mathbf{1}_i) \label{eq:slot_start_obj}
\end{eqnarray}
where $P^h_i$ is the attention probability distribution corresponding to the token $x_i,$ head $h$ of the final transformer layer, $\mathbf{1}_i$ is the indicator distribution over all input tokens with $1$ at position $i,$ and $0$ elsewhere, and $N_1, N_2$ are normalizing constants. The slot pair loss $\mathcal{L}_{\operatorname{slot-pair}}$ encourages the attention distribution for two slot words $x_i, x_j$ with different slot labels $y_i,y_j$ to focus on a disjoint set of context words. The second non-degenerate loss $\mathcal{L}_{\operatorname{non-deg}}$ prevents the slot pair loss to converge to a degenerate solution where each token mainly focuses on itself.  The final compositional loss for training our SLU model is given by:
\begin{equation}
\mathcal{L} = \mathcal{L}_{\operatorname{intent}} + \lambda_1 \mathcal{L}_{\operatorname{slot}} - \lambda_2 \mathcal{L}_{\operatorname{slot-pair}} - \lambda_3 \mathcal{L}_{\operatorname{non-deg}} \label{eq:comp_slu_loss}
\end{equation}
where $\lambda_1, \lambda_2, \lambda_3$ are hyper-parameters.

\subsubsection{Paired Training to Improve Length Generalization:}
 
The compositional loss enables the SLU model to better generalize to utterances with new combination of slots not seen during training (Section \ref{sec:experiment}). However, the model still perform poorly in {\em length generalization} splits. Length generalization has been shown to be particularly difficult for both sequence generation \cite{SCANLakeB:18, EkinAAA:20}, and multimodal tasks \cite{GaoHM:20}. We hypothesize that the model fails to generate good hidden state representations when presented an utterance with many slots, greater than those learned at training. To mitigate this problem, we develop an effective data augmentation approach we refer as {\em paired training}. Previously, a data augmentation approach GECA has been used to improve compositional generalization in sequence-to-sequence models \cite{SCANLakeB:18}. However, their performance on length generalization remained poor since it only replaces words/phrases in existing training sentences, and does not necessarily produce very long sentences. In our approach, we randomly select two distinct training utterances of the same intent but a disjoint combination of slots, and concatenate them with a period separator to form a new training sample. This exposes the model both to longer sequences, as well as new combination of slots not present in the original training set, resulting in better length generalization. Note that, neuro-symbolic approaches have been shown to perform effective length generalization in seq-to-seq models \cite{NyeS0L:20,ChenLYSZ:20}. However, these models are difficult to train, and they do not scale to real-world datasets which don't follow strict grammar rules. Model based data augmentation approach has also been explored for improving robustness of SLU models \cite{ZhaoZY:19}. However, this requires additional NL template information for each intent/slot which is difficult to obtain for large domains (e.g. ATIS).

\section{Experiments} \label{sec:experiment}

In this section we discuss our numerical results.

\setlength{\tabcolsep}{0.15em}
\begin{table*}[htbp]\scriptsize
  \centering
  \caption{Performance on Our Compositional Splits. The (train, test) sizes of Novel Slot Combination Split are (1229, 496) in ATIS and (1939, 600) in SNIPS. Sizes of Length Generalization Split are (1494, 163) in ATIS and (7107, 253) in SNIPS.}
    \begin{tabular}{p{1.1in}|cc|cc|cc|cc}
    \toprule
    \multicolumn{1}{c|}{\multirow{3}[4]{*}{\textbf{Model \textbackslash{} Dataset}}} & \multicolumn{4}{c|}{\textbf{ATIS}} & \multicolumn{4}{c}{\textbf{SNIPS}} \\
\cmidrule{2-9}          & \multicolumn{2}{c|}{\textbf{Novel Slot Combination}} & \multicolumn{2}{c|}{\textbf{Length Generalization}} & \multicolumn{2}{c|}{\textbf{Novel Slot Combination}} & \multicolumn{2}{c}{\textbf{Length Generalization}} \\
          & Slot (F1) & \multicolumn{1}{c|}{Intent (acc)} & Slot (F1) & Intent (acc) & Slot (F1) & Intent (acc) & Slot (F1) & Intent (acc) \\
    \midrule
    Full BERT SLU (upper bound) & 98.83  $\pm$ 0.05  & 98.39 $\pm$ 0.02 & 97.97  $\pm$ 0.07  & 99.39  $\pm$ 0.03  & 97.17  $\pm$ 0.05  & 98.67  $\pm$ 0.02  & 97.51  $\pm$ 0.07  & 99.62  $\pm$ 0.02  \\
    \midrule
    BERT SLU & 97.46 $\pm$ 0.05 & 97.72 $\pm$ 0.09 & 90.30 $\pm$ 0.25 & 96.32 $\pm$ 0.02 & 94.11 $\pm$ 0.27 & 97.17 $\pm$ 0.08 & 92.29 $\pm$ 0.69 & 99.13 $\pm$ 0.46 \\
    BERT SLU + relative pos emb & 94.83 $\pm$ 0.06 & 95.16 $\pm$ 0.15 & 91.69 $\pm$ 0.18 & 93.87 $\pm$ 0.03 & 94.27 $\pm$ 0.15 & 96.00 $\pm$ 0.08 & 89.77 $\pm$ 0.51 & 98.85 $\pm$ 0.12 \\
    BERT SLU + parse tree & 97.64 $\pm$ 0.06 & 96.57 $\pm$ 0.03 & 92.27 $\pm$ 0.12 & 96.07 $\pm$ 0.30 & 94.12 $\pm$ 0.17 & \textbf{98.33} $\pm$ 0.02 & 92.36 $\pm$ 0.21 & 98.82 $\pm$ 0.03 \\
    \midrule
    Comp. SLU (Ours) & $\textbf{98.10}^{\dagger}$ $\pm$ 0.12 & \textbf{97.78} $\pm$ 0.41 & $\textbf{94.93}^{\dagger}$ $\pm$ 0.68 & $96.93^{\dagger}$ $\pm$ 0.30 & $\textbf{95.37}^{\dagger}$ $\pm$ 0.42 & $97.83^{\dagger}$ $\pm$ 0.59 & $\textbf{95.63}^{\dagger}$ $\pm$ 0.92 & $\textbf{99.64}^{\dagger}$ $\pm$ 0.11 \\
     - Comp. Loss & 97.81 $\pm$ 0.14 & 96.97 $\pm$ 0.03 & 94.25 $\pm$ 0.12 & 96.87 $\pm$ 0.18 & 95.05 $\pm$ 0.10 & 97.33 $\pm$ 0.10 & 95.61 $\pm$ 0.85 & 99.60 $\pm$ 0.03 \\
     - Paired Training & 97.78 $\pm$ 0.04 & 96.98 $\pm$ 0.06  & 91.82 $\pm$ 0.13 & \textbf{97.79} $\pm$ 0.56 & 95.13 $\pm$ 0.13 & 96.17 $\pm$ 0.07 & 92.11 $\pm$ 0.32 & 98.81 $\pm$ 0.02 \\
    \bottomrule
    \end{tabular}%
  \label{tab:comp_results}%
  \begin{tablenotes}
     \item[1] * Full BERT SLU is trained using the whole standard training dataset, which indicates the performance upper bound. \item[2] $\dagger$ implies a significant improvement (p-value $<0.05$) using t-test over baseline BERT SLU model.
  \end{tablenotes}
\end{table*}

\subsection{Settings and baselines} \label{sec:params}

We train SLU models jointly for intent classification, and slot tagging tasks. Our evaluation metrics are {\em slot tagging F1 score}, and {\em intent accuracy}. Incorporating dependency parse information is known to improve compositional generalization of neural networks \cite{CirikBM:18,KuoKB:20}. We test an advanced baseline model ({\bf BERT SLU + parse tree}) which modifies the original attention scores in the final transformer layer with a weight inversely dependent on token distance on dependency tree. Intuitively, tokens which are further away in the dependency tree are assumed to be less informative, and given lower attention scores. We also test a third baseline ({\bf BERT SLU + relative pos emb}) which incorporates relative position embedding in BERT's attention computation \cite{HuangLXX:20}. In seq-to-seq tasks, it has been shown that relative position embedding helps in length generalization \cite{Ontanon:21}.

\noindent{\bf Parameters:} Our baseline and compositional SLU models are fine-tuned from BERT model {\em bert-base-uncased} \cite{BERTDevlinCLT:19}. We use hyper-parameters: batch size $32,$ learning rate $ \in \{10, 5\}\times 10^{-5},$ number of training steps $N \in \{4K, 5K, 6K\},$ $\lambda_1=1.$ For compositional models we use $\lambda_2=0.01, \lambda_3=0.1.$


\subsection{Results} \label{sec:unique}

\subsubsection{Results on novel slot combination split:}

First, we compare the performance of SLU models on the {\em novel slot combination} splits, where the test utterances have a distinct combination of slot types which doesn't appear together in training. Table \ref{tab:comp_results} presents the results (averaged over $5$ runs with different seeds). The first row corresponds to the performance of BERT SLU model when trained on the {\em full} standard training set $T_{\operatorname{train}}.$ This acts as an {\em upper bound} for the model performance. When the models are trained on the smaller compositional training set, the performance of the baseline models drop since they do not generalize well to the test sets. Observe that, in SNIPS compositional test set, the baseline performance drops about $3\%$ F1 score. The intent accuracy also drop around $1\%.$ BERT combined with dependency parse tree, fails to improve slot tagging performance, but it improves the intent accuracy. BERT with relative position embedding improves slot tagging slightly on SNIPS, but has poor intent accuracy. In contrast, our compositional SLU model improves slot tagging performance around $1\%$ over baseline while maintaining similar intent accuracy. In ATIS, we observe $1\%$ drop in F1 score of baseline model in test set. Our compositional model still improves performance over baseline. Note that, most models seem to perform better compositional generalization in ATIS split, than on SNIPS split. This is because, although both ATIS and SNIPS test sets have similar number of slot combinations ($210$ in ATIS, and $201$ in SNIPS), in the compositional training set ATIS has much larger number of distinct combinations $661,$ versus only $406$ for SNIPS. This helps models in ATIS learn better compositionality.


\subsubsection{Results on length generalization split:}

Next, we investigate the ability of SLU models to perform length generalization. For this experiment we consider the split with maximum two slots per training utterance. Table \ref{tab:comp_results} compares the model performance for SNIPS, and ATIS splits. Recall that, the top row indicates an upper bound on performance when the baseline model is trained on {\em full} training set $T_{\operatorname{train}}.$ We observe that when trained on compositional training set, the baseline model's slot tagging performance drops significantly; $5\%$ for SNIPS, and $7\%$ for ATIS. The intent accuracy suffer $3\%$ degradation in ATIS, and $1\%$ in SNIPS. BERT with dependency parse information show a small improvement. BERT with relative position embedding improves slot tagging in ATIS, but not in SNIPS indicating poor generalization across datasets. In contrast, our compositional SLU model significantly improves slot tagging performance, with about $5\%$ F1 score in ATIS and $4\%$ F1 score in SNIPS. The models also achieve similar intent accuracy as baseline. We further analyze the distribution of F1 scores w.r.t. the number of slots per test utterance in Table \ref{tab:length_err_analysis}. We observe our compositional model consistently improves F1 score over baseline model, irrespective of the number of slots/ utterance.

\setlength{\tabcolsep}{0.9em}
\begin{table}[t]\scriptsize
  \centering
  \caption{Slot Error Analysis in Length Generalization Split. $L$ denotes the number of slots/utterance in test split}
    \begin{tabular}{l|c|c|c|c}
    \toprule
    \multicolumn{1}{c|}{\multirow{2}[2]{*}{\textbf{L}}} & \multicolumn{2}{c|}{\textbf{ATIS (F1 score)}} & \multicolumn{2}{c}{\textbf{SNIPS (F1 score)}} \\
          & BERT SLU & Comp. SLU & BERT SLU & Comp. SLU \\
    \midrule
    $2$ & $76.0$ & $\mathbf{88.46}$ & $100.0$ & $100.0$ \\
    $3$ & $89.35$ & $\mathbf{94.01}$ & $93.26$ & $\mathbf{95.43}$ \\
    $4$ & $93.56$ & $\mathbf{95.58}$ & $93.56$ & $\mathbf{95.05}$ \\
    $5$ & $91.34$ & $\mathbf{92.19}$ & $93.02$ & $\mathbf{97.78}$ \\
    $6$ & $95.77$ & $95.77$ & $91.49$ & $\mathbf{96.91}$ \\
    $7$ & $92.86$ & $\mathbf{100.0}$  & N/A & N/A \\
    \bottomrule
    \end{tabular}
  \label{tab:length_err_analysis}
  \begin{tablenotes}
     \item[1] \scriptsize{* In our SNIPS test set there are no utterances with more than $6$ slots.}
  \end{tablenotes}
\end{table}


\subsubsection{Results on standard split:}

We also train our compositional model on full standard training set $T_{\operatorname{train}},$ and evaluate on standard test set $T_{\operatorname{test}}$ for both ATIS and SNIPS. In Table \ref{tab:full}, we compare the performance with baseline BERT SLU trained with same hyper-parameters. We observe that the compositional model achieves similar or better accuracy and F1 scores than baseline model in both datasets. Recall that, the standard train/test split in these benchmark datasets were generated randomly and hence they have a similar distribution. So it is expected that the BERT SLU model can have comparable performance to a compositional SLU model. However, as we discussed in Section \ref{sec:intro}, in real world cold-start settings this is often not the case which necessitates our compositional SLU model.

\setlength{\tabcolsep}{0.2em}
\begin{table}[t]\scriptsize
  \centering
  \caption{Performance on Standard Splits}
    \begin{tabular}{l|cc|cc}
    \toprule
    \multicolumn{1}{c|}{\multirow{2}[2]{*}{\textbf{Model \textbackslash{} Dataset}}} & \multicolumn{2}{c|}{\textbf{ATIS}} & \multicolumn{2}{c}{\textbf{SNIPS}} \\
          & Slot (F1) & Intent (acc) & Slot (F1) & Intent (acc) \\
    \midrule
    BERT SLU & 98.25 $\pm$ 0.02 & 97.8 $\pm$ 0.03  & 96.57 $\pm$ 0.16 & 98.97 $\pm$ 0.02 \\
    Comp. SLU & $\textbf{98.42}^{\dagger}$ $\pm$ 0.09 & $\textbf{98.2}^{\dagger}$ $\pm$ 0.02 & $\textbf{96.85}^{\dagger}$ $\pm$ 0.09 & $\textbf{99.11}^{\dagger}$ $\pm$ 0.07 \\
    \bottomrule
    \end{tabular}
  \label{tab:full}
  \begin{tablenotes}
     \item[1] \scriptsize{* For this experiment on SNIPS, we use the bert-base-cased model which has a slightly better performance. $\dagger$ implies a significant improvement (p-value  $<0.05$) using t-test over baseline BERT SLU model.}
  \end{tablenotes}
\end{table}


\subsubsection{Ablation study:}

Finally, we perform an ablation study to better understand the contribution of each component of our compositional SLU model. The two bottom rows in Table \ref{tab:comp_results} show the performance of our model when individual components (a) compositional loss, and (b) paired training are removed. We observe that for novel slot combination split, the model suffers similar drop in F1 score in test set, when the above two components are removed. This indicates they have a similar effect on the model for this split. However, in length generalization split of both datasets, the compositional model suffer significant drop (around $3\%$) in F1 score when paired training is removed, but suffer smaller degradation by removing the compositional objective. This supports our hypothesis that paired training plays a significant effect in length generalization. 


\section{Related work} \label{sec:related}

In spoken language understanding and dialog systems, the semantic meaning of an utterance is often represented in terms of intents, and slots \cite{WangDengAce:05,RayRic:07,YamDenYuWang:08,TurHakHil:11,HecHak:12}. Deep recurrent neural networks trained on labeled datasets have been shown to be very effective at jointly parsing an utterance into such intent and slot labels \cite{HakTurCelChenGao:16,LiuLane:16,KimLeeStratos:17,GooGaoHsuHuo:18,WangShenJin:18}. 

Recently, large scale pre-trained language models based on transformer architecture \cite{TransformerVaswaniSPUJGKP:17} have been widely successful for a variety of natural language understanding tasks \cite{BERTDevlinCLT:19,Roberta:19,XLNetYangDYCSL:19,DeBERTa:20}. The pre-trained model BERT when fine-tuned on joint intent classification and slot tagging task has achieved state-of-the-art performance on benchmark datasets \cite{BERTSLUChenZhuoWanh:19}. These SLU models however often suffer from significant performance drop in presence of open vocabulary slots \cite{RayShenJin:19,YanHXLMHX:20}. 

Traditional neural network models lack the ability to perform compositional generalization which humans can do easily \cite{fodor1988connectionism}. Recently, this phenomenon was demonstrated in seq-to-seq models for semantic parsing task \cite{SCANLakeB:18, Hupkes:20,ShawCPT:21,HerzigB20}. Many approaches have been explored to improve this issue using data augmentation \cite{Andreas:20,EkinAAA:20}, equivariant models \cite{GordonLBB:20}, meta-learning \cite{Lake:19}, and neuro-symbolic approaches \cite{NyeS0L:20,ChenLYSZ:20}. However, these techniques do not scale well to real world datasets. 

Compositional generalization has also been explored recently in multimodal setting for tasks such as robot navigation \cite{GSCANRuisABBL:20,HillLSCBMS:20}, VQA \cite{GQAHudsonM:19}, and so on. Models using dependency parse information \cite{CirikBM:18,KuoKB:20}, graph based reasoning \cite{HudsonM:19,SaqurN:20,GaoHM:20}, and multi-task learning \cite{HeinzeDeml2020ThinkBY} have improved compositionality of neural network models. In this paper, we explore compositional generalization of SLU models based on transformer architecture, trained jointly for intent classification, and slot tagging tasks.       


In this paper, we explore compositional generalization of SLU models based on transformer architecture, trained jointly for intent classification, and slot tagging tasks. 

\section{Conclusion} \label{sec:conclusion}

In this work, we demonstrate that SOTA SLU models based on pre-trained language models have poor generalization when: (1) an utterance has a novel combination of slots unseen during training, and (2) when an utterance has more slots than in any training utterances; scenarios which the models often encounter in practice. We develop a new compositional SLU model to tackle these issues. First showing that, by adding a new compositional loss, the model's attention distribution can better focus on informative words, thereby improving model's generalization to novel slot combination. We further propose a new paired training data augmentation technique which greatly improves length generalization. In our future work, we want to further explore the impact of OOV slot values on compositionality.

\bibliographystyle{IEEEtran}
\bibliography{compositional_slu}

\end{document}